\def\year{2021}\relax
\documentclass[letterpaper]{article} 
\usepackage{aaai21}  
\usepackage{times}  
\usepackage{helvet} 
\usepackage{courier}  
\usepackage[hyphens]{url}  
\usepackage{graphicx} 
\usepackage{natbib}  
\usepackage{caption} 
\usepackage{subfigure}
\usepackage{verbatim}
\usepackage{booktabs}
\usepackage{setspace}
\usepackage{amsmath}
\usepackage{amsthm}
\usepackage{amssymb}
\usepackage{algorithm}
\usepackage{algorithmic}
\usepackage{multirow}

\begin{document}
\title{Disentangled Neural Architecture Search}

\author {
        Xinyue Zheng \textsuperscript{$\dagger$},     
        Peng Wang \textsuperscript{$\dagger$}\thanks{Corresponding Author},
        Qigang Wang \textsuperscript{$\dagger$}.
        Zhongchao Shi \textsuperscript{$\dagger$} \\
        \textsuperscript{$\dagger$}AI Lab, Lenovo Research, Beijing, 100089, China\\
zhengxy10@lenovo.com, wangpeng31@lenovo.com, wangqg1@lenovo.com, shizc2@lenovo.com\\}

\maketitle

\renewcommand{\thefootnote}{\fnsymbol{footnote}}

\begin{abstract}

\begin{quote}

Neural architecture search has shown its great potential in various areas recently. However, existing methods rely heavily on a black-box controller to search architectures, which suffers from the serious problem of lacking interpretability. In this paper, we propose disentangled neural architecture search (DNAS) which disentangles the hidden representation of the controller into semantically meaningful concepts, making the neural architecture search process interpretable. Based on systematical study, we discover the correlation between network architecture and its performance, and propose a dense-sampling strategy to conduct a targeted search in promising regions that may generate well-performing architectures. We show that: 1) DNAS successfully disentangles the architecture representations, including operation selection, skip connections, and number of layers. 2) Benefiting from interpretability, DNAS can find excellent architectures under different FLOPS restrictions flexibly. 3) Dense-sampling leads to neural architecture search with higher efficiency and better performance. On the NASBench-101 dataset, DNAS achieves state-of-the-art performance of $94.21\%$ using less than $1/13$ computational cost of baseline methods. On ImageNet dataset, DNAS discovers the competitive architectures that achieves $22.7\%$ test error. our method provides a new perspective of understanding neural architecture search.

\end{quote}
\end{abstract}

\section{Introduction}

Neural architecture search (NAS) aims to automatically generate network architectures as the substitute of manual design and has been widely explored in various areas, such as computer vision \cite{TransferableNAS,HieraRepre}, language modeling \cite{DARTS,ENAS} and model compression \cite{he2018amc,dong2019network}. Recent NAS approaches typically employ various optimization methods (e.g. evolutionary algorithm \cite{2018Regularized,real2017large}, differentiable optimization \cite{DARTS,Liang2019DARTS,NAO}, and reinforcement learning \cite{ENAS,TransferableNAS} to achieve impressive performances through complex end-to-end training.

However, recent NAS methods usually design a large and expressive search space for maximal performance, which will result in very expensive training and search processes. What's more, black-box optimization of controller makes human can only use evaluation criteria (e.g. accuracy and error rate) to assess the credibility of the algorithm. Since the computational bottleneck of NAS is to aimlessly find the best architecture from the huge search space, making the search process interpretable \cite{ExplanationAI} may help improve the algorithm efficiency. Interpretability enables every decision of the controller transparent to human, thus ensuring that people can understand which architecture signature contributes to better performance and reducing the large search space into a smaller but promising region.

\begin{figure}[!tp]
\centering
\includegraphics[scale=0.27]{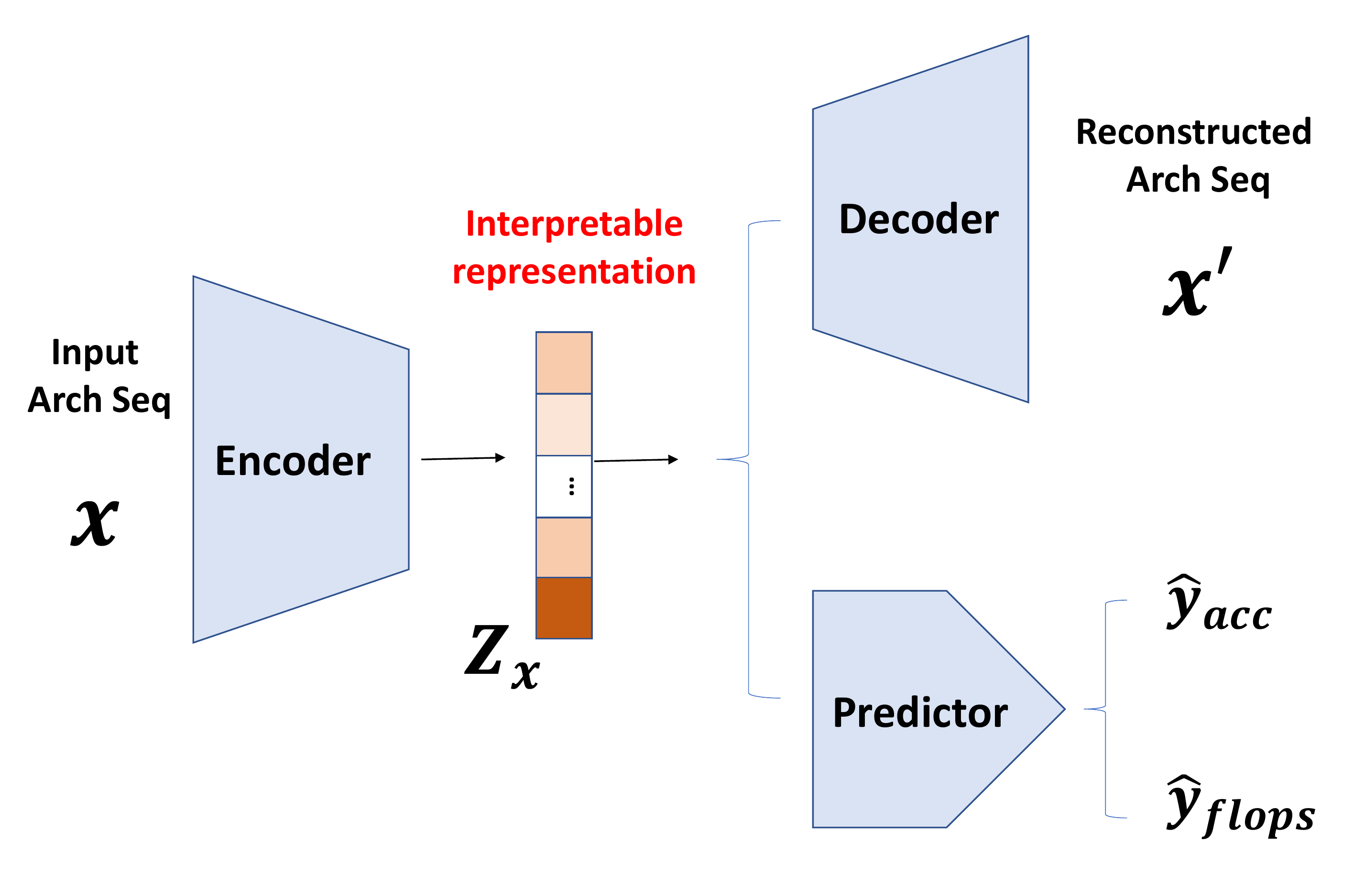}
\caption{The overall framework of DNAS. It disentangles the embedding of input architecture $x$ into interpretable representations $z_x$, and $z_x$ can be optimized via maximizing the predictive accuracy $\hat{y}_{acc}$ and minimizing the predictive FLOPS $\hat{y}_{flops}$. The decoder can reconstruct the architecture sequence from $z_x$.}
\end{figure}

In this paper, we rethink the process of neural architecture search by investigating the following open questions: Is there any interpretable disentangled factors that can control the independent representation of the architecture? Is there a systematic connection between these disentangled factors and their corresponding network performance? Do well-performing neural networks have common structural characteristics? If the answer to all the above questions is 'yes', does analyzing the distribution of disentangled factors contributes to quickly locate the search space of good architectures, hence improving the efficiency of NAS algorithm?

To learn interpretable representations, some studies \cite{2017Disentangled,2019Diagnosing,2018Understanding} based on generative models learn to produce real data from Gaussian distribution, where the convexity of the Gaussian density makes linear operations between representations meaningful \cite{2020A}. In our work, we leverage $\beta$-VAE to disentangle the representation of architecture, and determine the potential search regions that may produce high-performance architectures through the distribution of the disentangled factors, then perform dense-sampling among these regions to find good architectures efficiently.

Specifically, we employ encoder-predictor-decoder framework proposed by NAO \cite{NAO} as the controller. Using similar training method of $\beta$-VAE, the encoder takes architecture sequence as input and learns to disentangle the information into human-understandable semantic concepts, such as operation conversion, skip connection, and number of layers. Taking the disentangled embedding as input, the predictor can predict the accuracy and FLOPS of given architectures, which ensures that a well-performing architecture is generated under certain FLOPS constraints. Importantly, interpretable semantic factors combined with accurate prediction of performance, can help determine the promising regions of disentangled factors for architecture's future improvements. During each sampling period, the embedding is dense sampled from the expected promising regions with a certain probability, and randomly generated from the entire search space to avoid omitting some good architectures. Based on the observation that FLOPS and accuracy are usually positively correlated, we also perform dense sampling in the edge area of the FLOPS constraints. We conduct the experiments under semi-supervised setting \cite{semiNAS} for reducing the number of architecture evaluations. Empirically, we show that DNAS achieves impressive performance on NASBench-101 and ImageNet classification datasets.

Our work makes the following contributions:

\begin{itemize}
\item We propose DNAS, an interpretable and efficient approach for NAS, which is able to achieve higher performance under fixed resource budgets.

\item Through disentangling the structural representations to human-understandable semantic concepts, we can intuitively understand the relation between architecture and its performance.

\item Locating promising regions of dense-sampling is simple yet effective, requiring only a few sample evaluations.

\item Dense-sampling encourages the generation of potentially good architectures, while alleviating the burden of aimlessly searching in a large search space.

\item DNAS achieves impressive results on computer vision tasks including NASBench-101 and ImageNet efficiently.

\end{itemize}

\section{Related Work}

\subsection{Interpretability of Neural Network}

The drawback of training the NAS black-box models is lacking of interpretability. To make hidden representations of neural network interpretable, some researches \cite{2018Interpretable,2020Architecture,survey,2017Interpretability} design specific loss function and network for mapping semantic concepts on the layers or paths of network to obtain an interpretable network. Other methods \cite{2017Disentangled, 2018Understanding,2016InfoGAN,2015Discovering,MGAN} adopts disentanglement of hidden representation to encourage learning interpretability of neural network. InfoGAN \cite{2016InfoGAN} learns to disentangle the representation by maximizing the mutual information between latent variables and its observation. $\beta$-VAE enforces the latent code to obey the standard Gaussian distribution, then attributes the semantic concepts to latent code by increasing the information capacity of the latent code. In computer vision, A face image can be disentangled into some semantic representations, e.g., posture, expression, color. Transferring to NAS field, we hope to obtain disentangled factors that are meaningful for neural network design.

\subsection{Neural Architecture Search}


Recent NAS methods , mainly based on evolutionary algorithms \cite{real2017large,2018Regularized}, reinforcement learning \cite{ENAS,2016Neural,TransferableNAS} and differentiable optimization \cite{DARTS,NAO,Liang2019DARTS} usually design a large and expressive search space for maximal performance, which will result in very expensive training and search processes. Traditional NAS methods \cite{2018Regularized,2016Neural,TransferableNAS} consume hundreds to thousands of GPU days to obtain fairly good architectures. To improve NAS efficiency, many studies propose targeted solution. ENAS \cite{ENAS} adopts weight sharing among different searched models which alleviates the burden of training every model from scratch. Single-Path NAS \cite{2020Single} reduces the number of trainable parameters by employing a single-path over-parameterized convolution network, which could express all the network structure with shared convolutional kernel parameters. NAO \cite{NAO} proposes an encoder-predictor-decoder framework of controller to map discrete space into continuous space to perform efficient optimization. Based on NAO framework, SemiNAS \cite{semiNAS} trains the controller in a semi-supervised manner, where a large number of unlabeled architectures (without evaluation) can be used to train the controller. Taking search efficiency as an important optimization goal, we also adopt the controller framework proposed by NAO and train it through semi-supervised learning. However, instead of randomly generating unlabeled architectures, our method is oriented to generate potentially excellent architectures to reduce search time.

\section{Approach}

In this section, we will firstly introduce the framework of our method. Then explain the approach of disentangling network architectures, and how to perform dense-sampling according to the distribution of disentangle factors.

\subsection{Framework}

We use the controller structure similar to NAO \cite{NAO}, but remove the attention mechanism to ensure completely disentanglement, and add a FLOPS predictor to make it feasible under resource constraints. As shown in Figure 1, the controller consists of three modules: encoder, decoder and predictor. The encoder $f_e$ maps an architecture sequence $x$ into hidden representation $z_x = f_{e}(x)$ through a single-layer LSTM. In order to make $z_x$ interpretable, we borrow the idea from $\beta$-VAE \cite{2018Understanding} which adopts an adjustable hyperparameter $\beta$ to encourage disentangling of latent representation $z_x$. The decoder $f_d$ is a multi-layer LSTM, responsible for decoding latent variable $z_x$ and reconstructing the architecture string $x'$, which can be formalized as $x'= f_d(z_x)$.

In practical engineering, optimizing accuracy is usually not the only target, we have to consider how many computational resources are available. Therefore, our predictor is designed for solving multi-objective optimization problem, which can predict both accuracy and FLOPS. Feed $z_x$ into the MLP predictor, we can obtain the predictive accuracy $f_{acc}(z_x)$ and FLOPS $f_{flops}(z_x)$ of the input architecture. Accurate performance prediction guides the controller to generate well-performing neural networks, while FLOPS prediction constraints the discovered architectures developed at a fixed computational budget.

\begin{figure}[!h]
\centering
\subfigure[]{
\begin{minipage}[t]{0.83\linewidth}
\centering
\includegraphics[width = 1\linewidth]{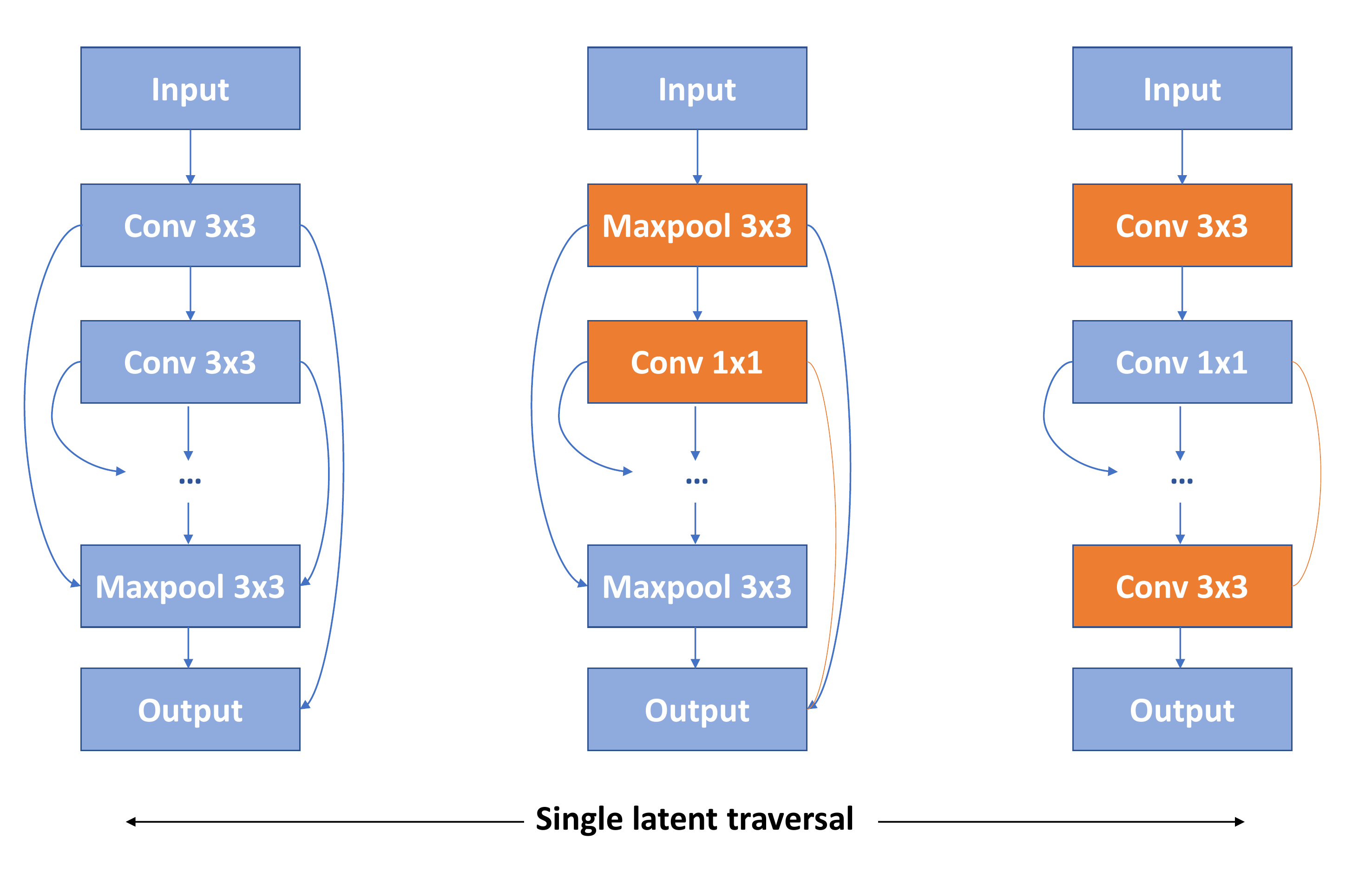}
\end{minipage}
}%

\hspace{.001in}
\subfigure[]{
\begin{minipage}[t]{0.84\linewidth}
\centering
\includegraphics[width = 1\linewidth]{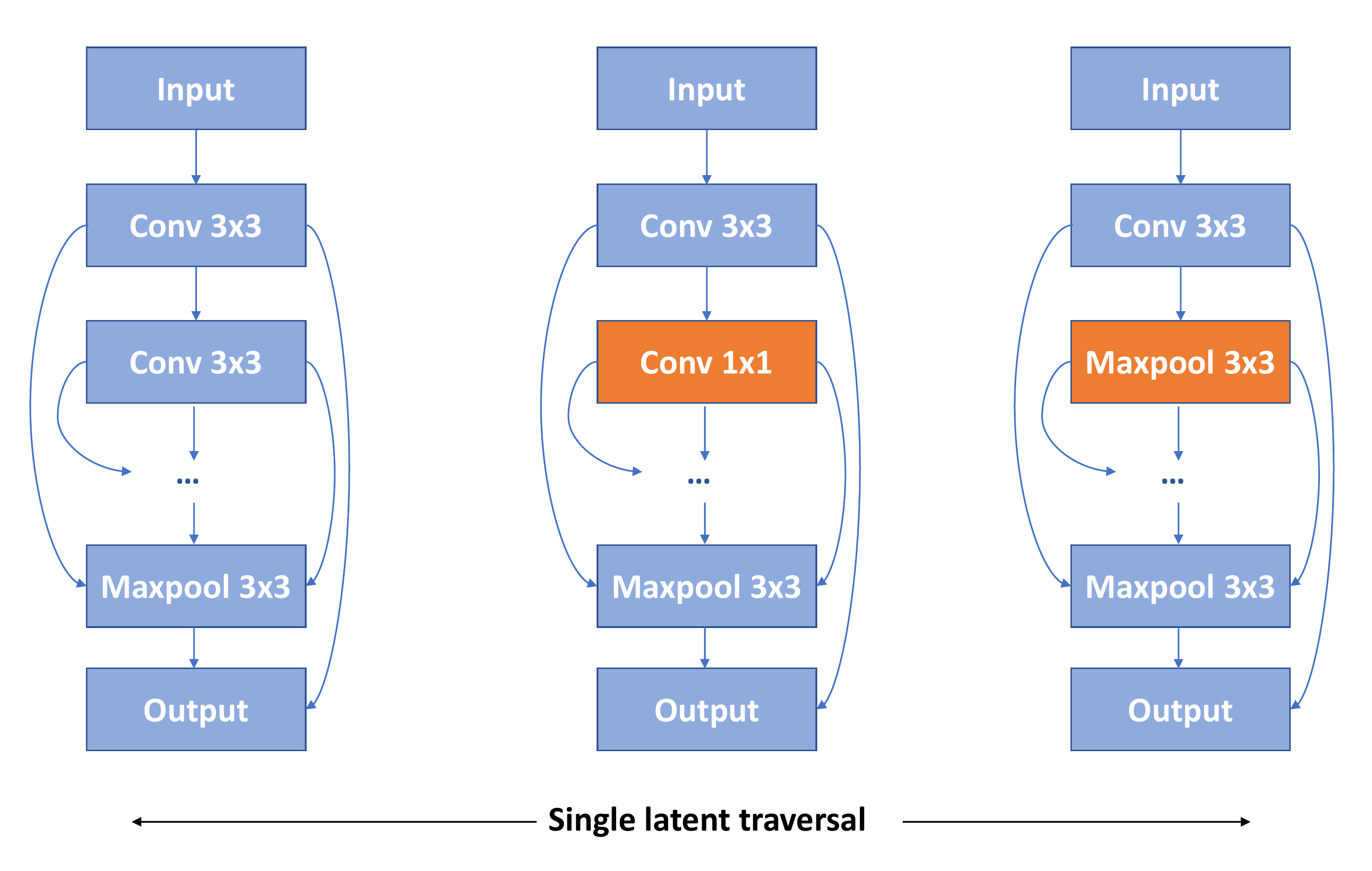}
\end{minipage}
}%
\hspace{.01in}
\centering
\caption{Entangled VS. disentangled representations of neural network structure. (a): The modification of a single entangle factor causes changes in multiple representations. (b): A disentangled factor only controls the operation conversion of the third layer.}
\end{figure}

\subsection{Disentangled Representation Learning}

To achieve controllable powerful architecture generation, semantically disentangled and interpretable factors of variation are indispensable. We leverage $\beta$-VAE to learn a disentangled latent space, where a single latent unit of $z_x$ is only sensitive to the change of single representation \cite{2012Representation}:

\begin{equation}
\begin{split}
L(\theta, \phi, x) & = L_{rec} + \beta L_{KL} \\
& = E_{q_\phi(z|x)}[logp_\theta(x|z)]-\beta D_{kl} (q_{\phi}(z|x)||p(z)),
\end{split}
\end{equation}

where the $\phi$ and $\theta$ is to parameterize the distribution of the encoder and the decoder. The first term aims to learn the marginal likelihood of the data in generative procedure, and the second term is Kullback-Leibler (KL) divergence between the true and the learned approximate posterior $q_{\phi}(z|x)$. It is suggested that finding an appropriate value of $\beta$ would be critical to disentangling latent representation. 

In our perspective, each well-disentangled factor could independently control different representations of an architecture, such as the operation of a certain layer, the number of layers of the architecture, the skip connection, and so on. Figure 2. shows an example of entangled versus disentangled representations of neural network structure.

\subsection{Efficient Dense-Sampling Strategy}

Inspired by \cite{2018MnasNet,2019EfficientNet}, we develop our DNAS method which simultaneously optimizes the accuracy and FLOPS to achieve multi-objective neural architecture search.

As described above, the encoder maps the discrete architecture sequence $x$ into continuous and interpretable representation $z_{x} = f_{e}(x)$, then employs the predictor to predict its corresponding accuracy $\hat{y}_{x}^{acc} = f_{acc}(z_x)$ and FLOPS $\hat{y}_{x}^{flops} = f_{flops}(z_x)$. The training of the predictor is to minimize the prediction loss of both accuracy and FLOPS:

\begin{equation}
L_{acc} = \sum_{x \in X} (y_{x}^{acc} - f_{acc}(z_{x}))^{2},
\end{equation}
\begin{equation}
L_{flops} = \sum_{x \in X} (y_{x}^{flops} - f_{flops}(z_{x}))^{2},
\end{equation}

where $X$ indicates all the candidate architecture sequences $x$ that have evaluated performances $y_{x}^{acc}$ and FLOPS information $y_{x}^{flops}$.

The combination of accurate performance prediction and disentangled semantic factors helps us understand which concepts affect the composition of high-accuracy network structures intuitively. Further, the obtained relation between accuracy and disentangled factors allows us to conduct a targeted dense-sampling in the hidden space where high-accuracy architectures are easier to be generated. Meanwhile, the prediction of FLOPS can ensure the generated architectures meet the constraints of computing resources.

We now introduce how to perform dense-sampling based on disentangled factors:

1) Randomly generate $M$ architectures $x_1, x_2, ..., x_M$ from the search space. Evaluating the performances of these architectures on validation set and calculating their FLOPS, and we can obtain the dataset $D' = \{(x_i, y_{x}^{acc}, y_{x}^{flops}), i = 1,2, ..., M\}$ to train the controller.

2) Use the trained encoder $f_e$ to map $x_{1,2, ..., M}$ into latent disentangled space $z_{1,2, ..., M}$. Suppose each latent code $z_i$ contains $S$ dimensions, for the $s$-$th$ dimension, we choose the $z$ values of the architectures with top-k accuracy as the promising region for dense-sampling:   
  
\begin{equation}
R_s = \underset{i \in I_k}{\cup}[z_i -\sigma, z_i+\sigma],
\end{equation}

where $I_k$ is the set of top-k architectures.

3) Sample from promising region $\mathbb{R}^{R_1, R_2, ..., R_S}$ with a probability of $\epsilon_1$. According to expert priors, well-performing neural networks are generally have deeper layers and higher FLOPS \cite{2016On,2014VeryDeep}, so we will also perform dense-sampling in the edge area that meets the FLOPS limit with $\epsilon_2$. To avoid missing good architecture, we sample from the entire search space with a probability of $\epsilon_3$, where $\epsilon_3 = 1 - \epsilon_1 - \epsilon_2$.

\subsection{Implementation of DNAS}

In this section, we will introduce how to use the disentanglement and dense-sampling to discover more expressive architectures. In general, the DNAS takes architecture-accuracy-FLOPS pairs as the training data to jointly train the encoder $f_e$, predictor $f_{acc}$ and $f_{flops}$, and decoder $f_d$ by minimizing the following loss function:

\begin{equation}
L_{total} = \alpha L_{acc}+ \lambda L_{flops} + \mu L_{rec} + \beta L_{kl},
\end{equation}

where the $L_{acc}$ and $L_{flops}$ are the prediction losses introduced in Eqn 2 and Eqn 3. $L_{rec}$ and $\beta L_{kl}$ are the architecture reconstruction loss and KL loss described in Eqn 1. The hyper-parameters $\alpha$, $\lambda$, $\mu$ and $\beta$ are used to trade off between these losses.

Our goal is to maximize the accuracy of the neural network architectures under any resource constraints, which can be formulated as:
\begin{equation}
\begin{split}
\text{maximize}  \text{  }  &  ACC(x),  \\
\text{subject to} \text{  } &  FLOPS(x) \leq F.
\end{split}
\end{equation}

Therefore, for architectures that meet FLOPS restrictions $FLOPS(x) \leq F + \tau $, where $\tau$ is the margin of FLOPS that may be optimized by the gradient descent, we perform the optimization process:

\begin{equation}
z_{x}^{'} = z_{x} + \eta_{1}\frac{\partial{f_{acc}(z_{x})}}{\partial{z_{x}}} - \eta_{2}\frac{\partial{f_{flops}(z_{x})}}{\partial{z_{x}}},
\end{equation}

where $\eta_1$ and $\eta_2$ is the step size. For any given disentangled factors $z_x$, DNAS moves its embedding $z_{x}$ towards the direction of the gradient ascent of the accuracy, while using gradient descent to reduce the FLOPS prediction. Finally, we feed the new representation $z_{x}^{'}$ into decoder to get an expected expressive architecture with better performance. The detailed algorithm of DNAS is shown in Algorithm 1. Note that we adopt semi-supervised setting proposed by \cite{semiNAS}, to accelerate search process by training the controller with a large number of unlabeled architectures (without evaluation).

\begin{algorithm}[!h]  
    \caption{Disentangled Neural Architecture Search under Semi-Superviseed Setting}  
    \begin{algorithmic}[1]  
        \STATE Randomly initialize $\phi$ for encoder, $\theta$ for decoder and $\vartheta$ for predictor 
        \STATE Randomly generate $M$ architectures. Train $T$ steps for each architecture.
        \STATE Evaluate $M$ architectures to obtain architecture-accuracy-flops pairs and form the labeled dataset $D'$.
    
        \FOR {$l$ $\leftarrow$ $1$ to $L$ do}
        
        \STATE Use labeled data in $D'$ to pre-train the controller.
        
        \STATE Analyze the distribution of disentangled factors, and select the high promising regions of $z$-value according to Eqn 4.

        \STATE Sample $N$ architectures, in which samples are taken from the promising region and FLOPS edge area with the probability of $\epsilon_1$ and $\epsilon_2$ respectively, and sample in the whole space with the probability of $\epsilon_3$

        \STATE Predict the accuracy and flops of sampled $N$ architectures using $f_{acc}$ and $f_{flops}$, and form dataset $\hat{D}$.
        
        \STATE Set $D = D' \cup \hat{D}$.
        
        \STATE Use the dataset $D$ to train controller by minimizing Eqn 5.
        
        \STATE Pick the top $P$ architectures that meet the FLOPS constraint and use gradient optimization in Eqn 7 to obtain better architectures and Integrate them into dataset $D'$.
        
        \ENDFOR 
    \end{algorithmic}  
\end{algorithm}

\section{Experiments}

In this section, considering a large number of candidate architectures that we want to explore, we first study the impact of modifying the disentangled factor $z_x$ on the semantic architecture on the NASBench-101 dataset \cite{bench101}. Then, we conduct further empirical research on both NASBench-101 and Imagenet \cite{2015ImageNet} benchmark datasets to evaluate the effectiveness and efficiency of DNAS methods.

\begin{figure*}[!h]
\centering
\subfigure[]{
\begin{minipage}[t]{0.41\linewidth}
\centering
\includegraphics[width = 1\linewidth]{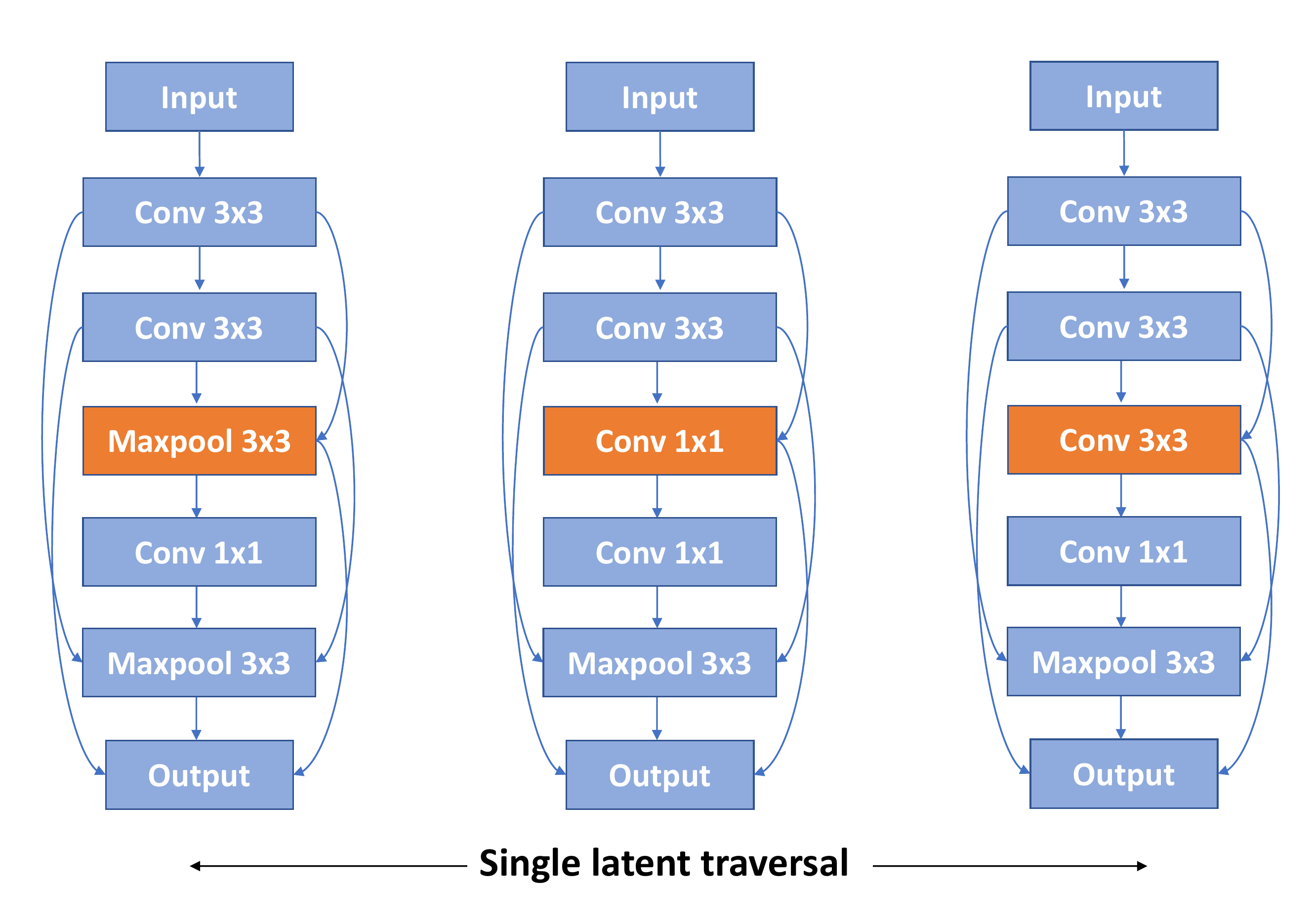}
\end{minipage}
}%
\hspace{.35in}
\subfigure[]{
\begin{minipage}[t]{0.50\linewidth}
\centering
\includegraphics[width =  1\linewidth]{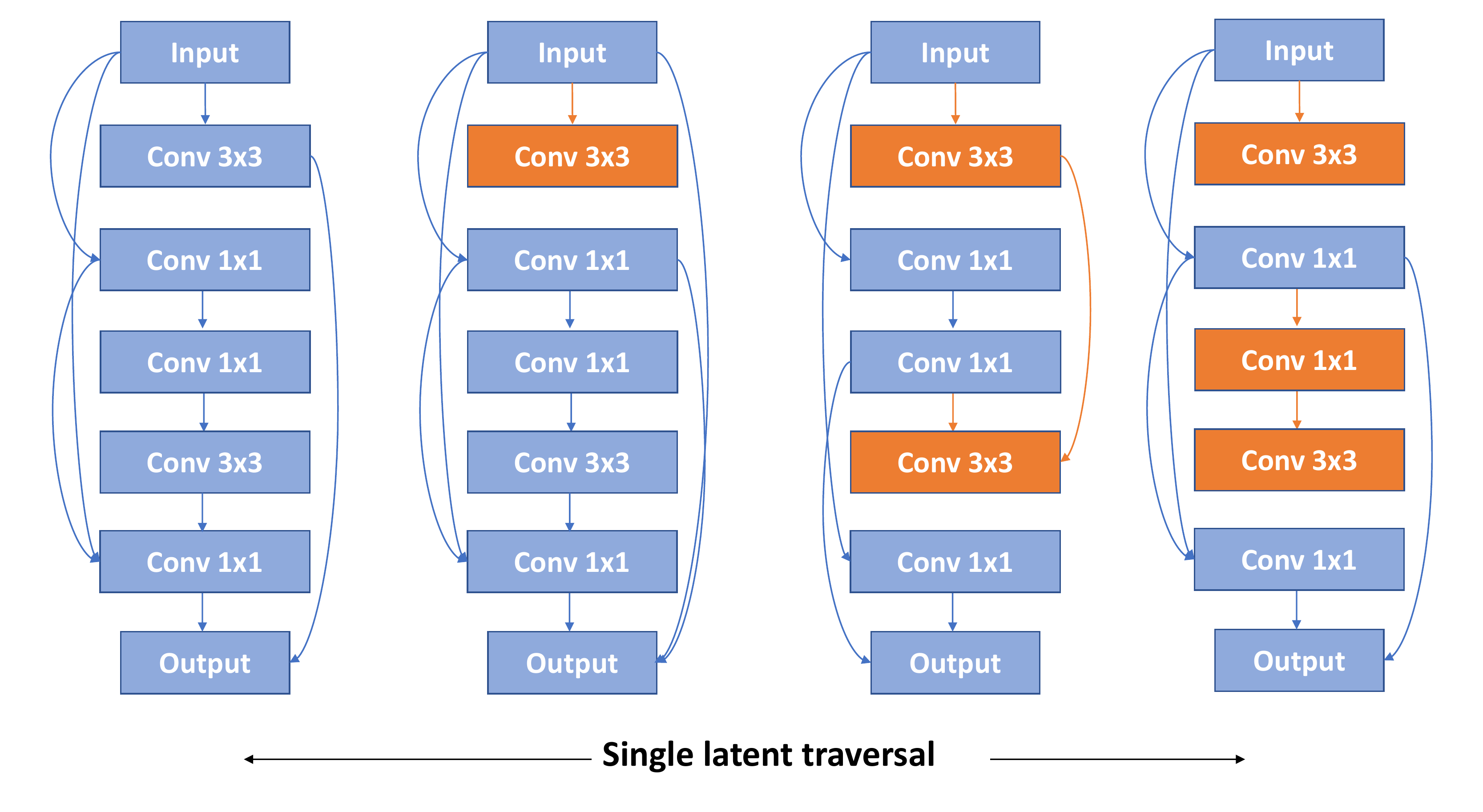}
\end{minipage}
}%
\hspace{.01in}
\centering
\caption{Disentangling and reconstructions from DNAS. The traversals of a single semantic factor result in smooth changes of the output architectures. (a): The $z_{24}$ latent factor independently codes for operation conversion. (b): The $z_{10}$ latent factor independently codes for the number of layers by controlling skip connection.}
\end{figure*}

\begin{figure*}[!h]
\centering
\subfigure[]{
\begin{minipage}[t]{0.22\linewidth}
\centering
\includegraphics[width = 1\linewidth]{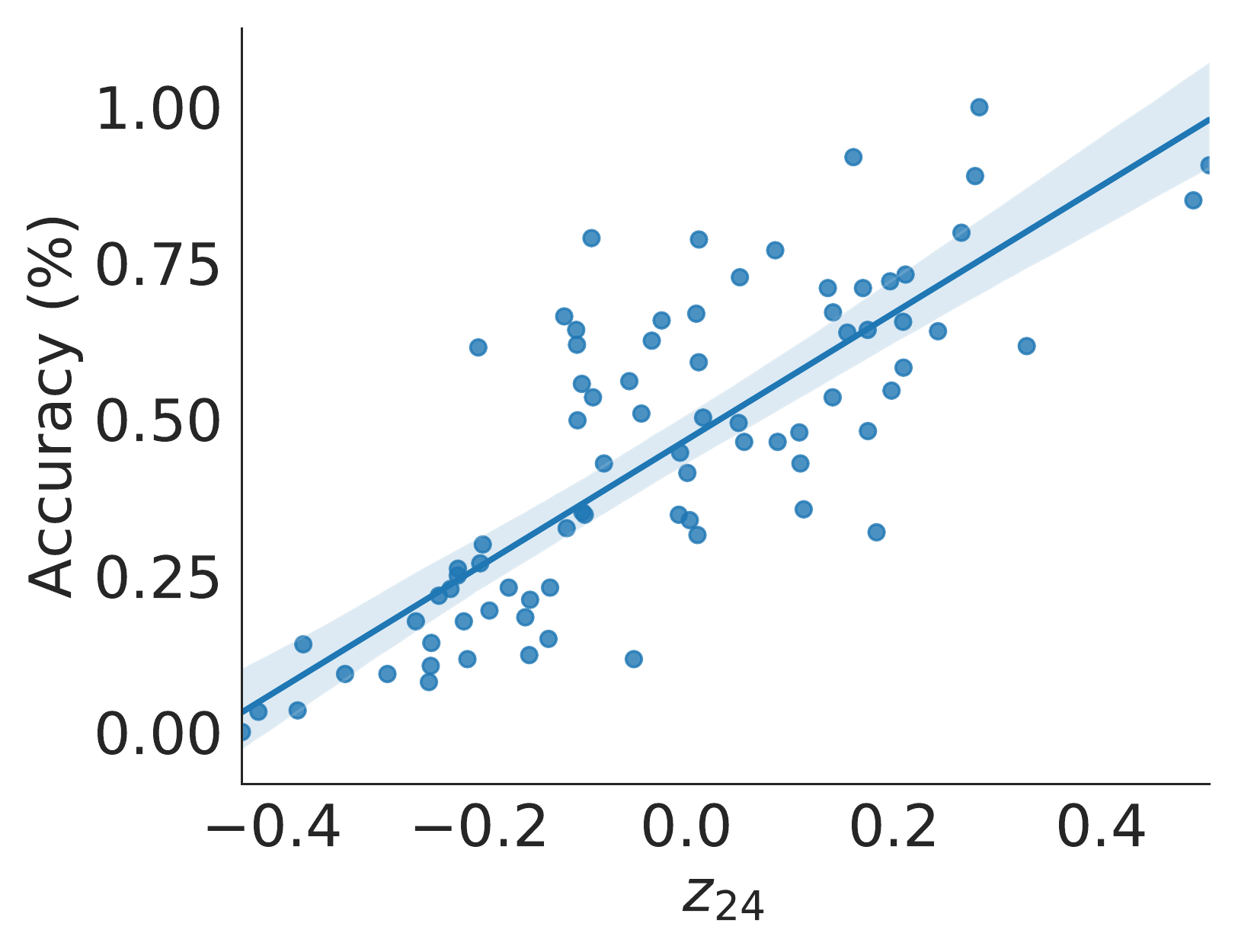}
\end{minipage}
}%
\hspace{.02in}
\subfigure[]{
\begin{minipage}[t]{0.23\linewidth}
\centering
\includegraphics[width =  1\linewidth]{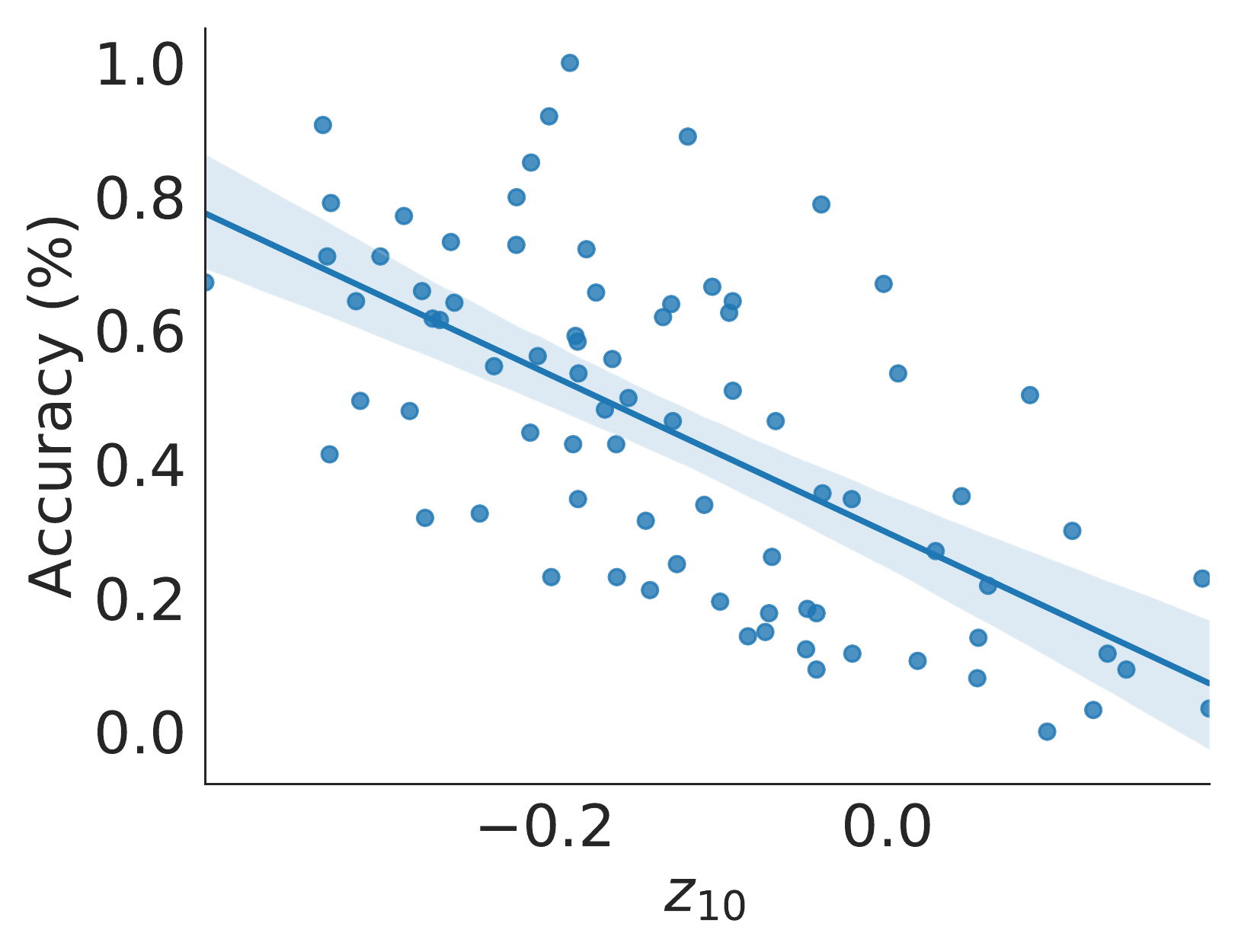}
\end{minipage}
}%
\hspace{.05in}
\subfigure[]{
\begin{minipage}[t]{0.23\linewidth}
\centering
\includegraphics[width =  1\linewidth]{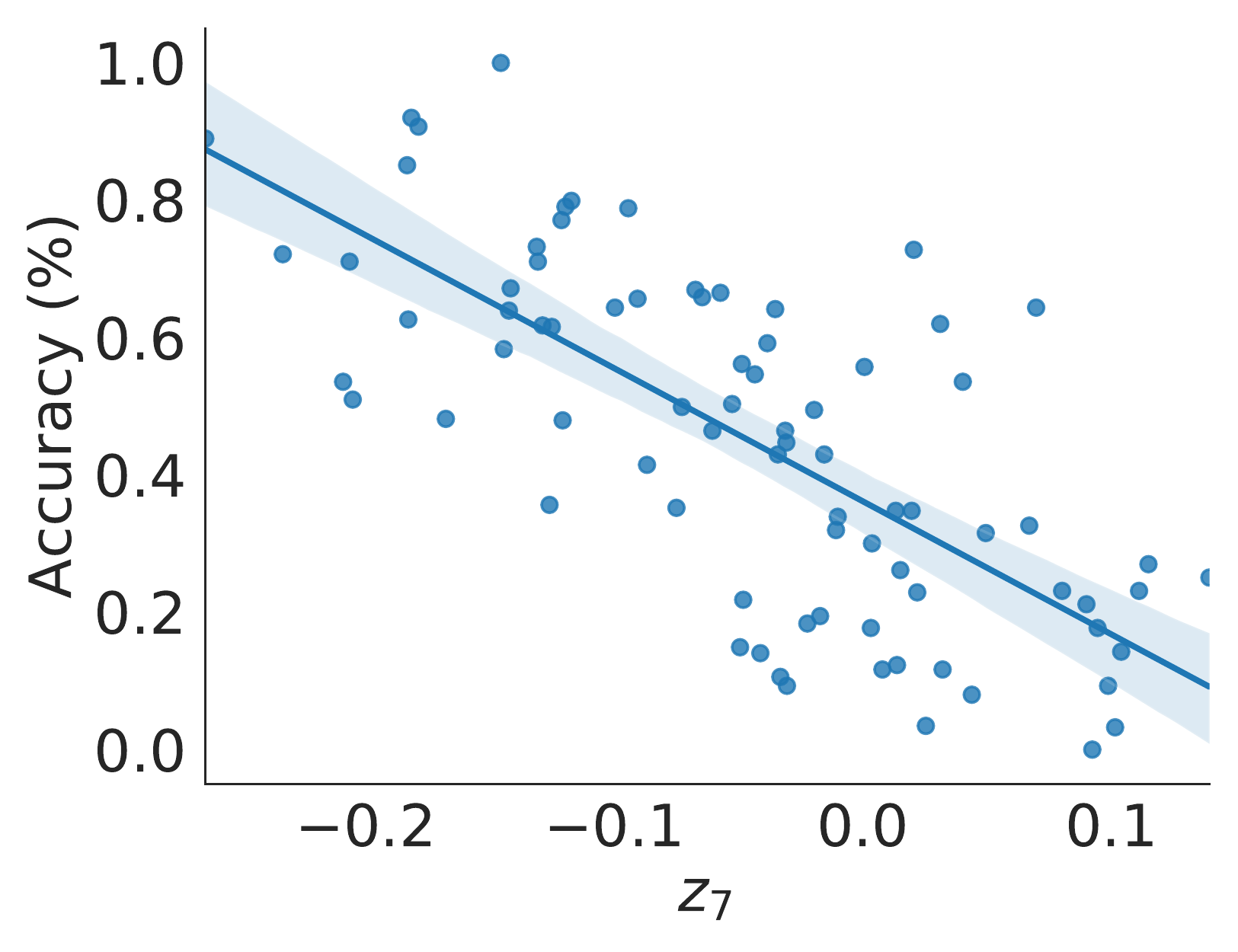}
\end{minipage}
}%
\hspace{.02in}
\subfigure[]{
\begin{minipage}[t]{0.23\linewidth}
\centering
\includegraphics[width =  1\linewidth]{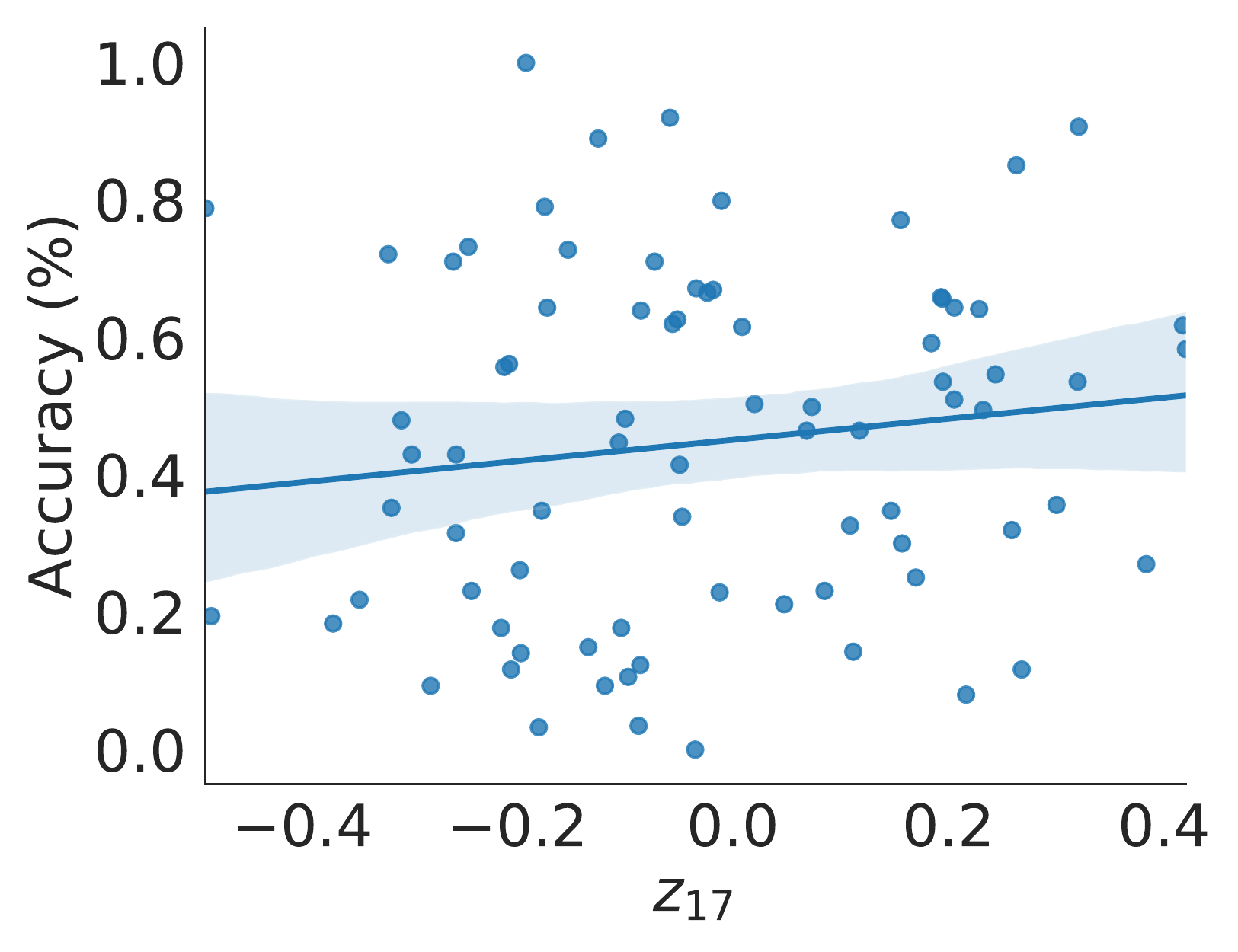}
\end{minipage}
}%

\hspace{.01in}
\centering
\caption{Correlation between disentangled factors and architectures performances on NASBench-101. We select four disentangled factors for display, and (a)-(c) have evident relation with accuracy.}
\end{figure*}

\subsection{Disentangling Architecture Representations}

\subsubsection{Dataset and Settings}

We conduct the disentangling experiments on NASbench-101 dataset. NASBench-101 is a new tabular benchmark for neural architecture search which designs a compact and expressive search space for CIFAR-10. It includes 423,624 unique architectures, and each network architecture is mapped to their training and evaluation metrics.

For disentangling experiments, we use single-layer LSTM with a hidden size of $26$ for both encoder and decoder and the accuracy predictor is a $3$-layer MLP with hidden units of $26, 64, 1$ respectively. The trade off in Eqn 5 is set as $\alpha = 0.8$, $\lambda = 0.3$, $\mu = 0.2$, and $\beta = 1$. For semi-supervised learning, we follow the \cite{semiNAS} setting, where N = 100 and M = 10000. Finally, we run the controller for $L = 2$ iterations with an initial learning rate of $0.001$.

\subsubsection{Results}

The controller of DNAS learns to reconstruct architectures from interpretable factors $z$, so we modify the value of $z$ to observe semantic architecture modification. The experimental results show that DNAS can successfully disentangle the human-understandable representation of the architecture, including operation conversion, skip connection and the number of layers. The same disentangle factor controls the same semantic concept in different network architectures.

For example, in Figure 3(a), we traverse the $24$th latent factor $z_{24}$ from $-0.4$ to $0.4$ for any given architecture, while keeping the remaining latent factor fixed, and reconstruct the architectures. The disentangling is obvious: only the operation of the third layer is changed. We can modify another semantic factor $z_{10}$ the same ways to obtain Figure 3(b).

To further quantify the disentangling, we investigate the relation between disentangled factor $z$ and architecture performance. Figure 4. demonstrates that there is a strong relation between disentangled factors and corresponding accuracy, which can infer that the specific latent factors are able to control the semantic representation of the architecture and ultimately improve the performance of the architecture. The disentangled factors of Figure 4(a) and (b) corresponds to those in Figure 3(a) and (b).

\begin{figure*}[!h]
\centering
\subfigure[]{
\begin{minipage}[t]{0.22\linewidth}
\centering
\includegraphics[width = 1\linewidth]{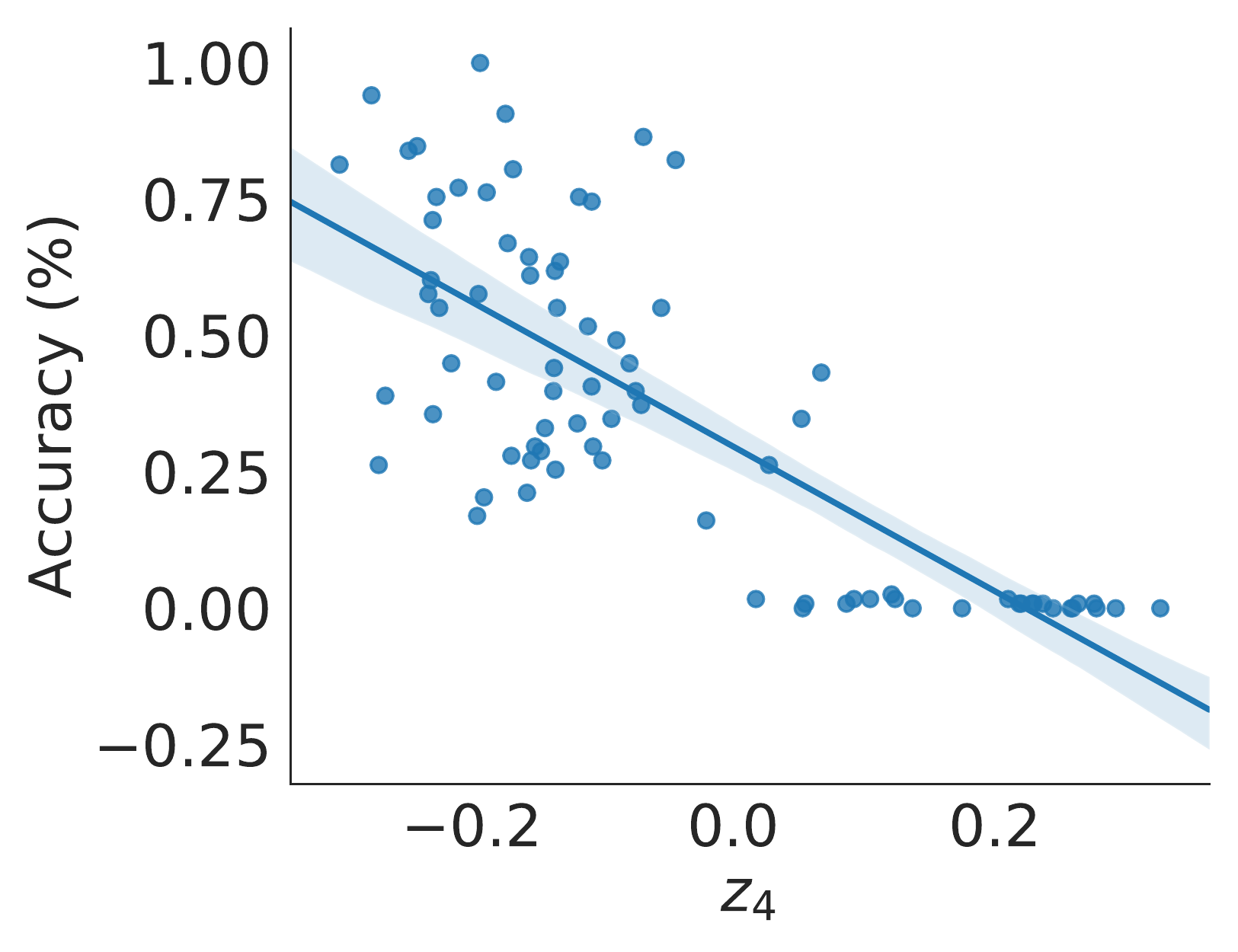}
\end{minipage}
}%
\hspace{.02in}
\subfigure[]{
\begin{minipage}[t]{0.23\linewidth}
\centering
\includegraphics[width =  1\linewidth]{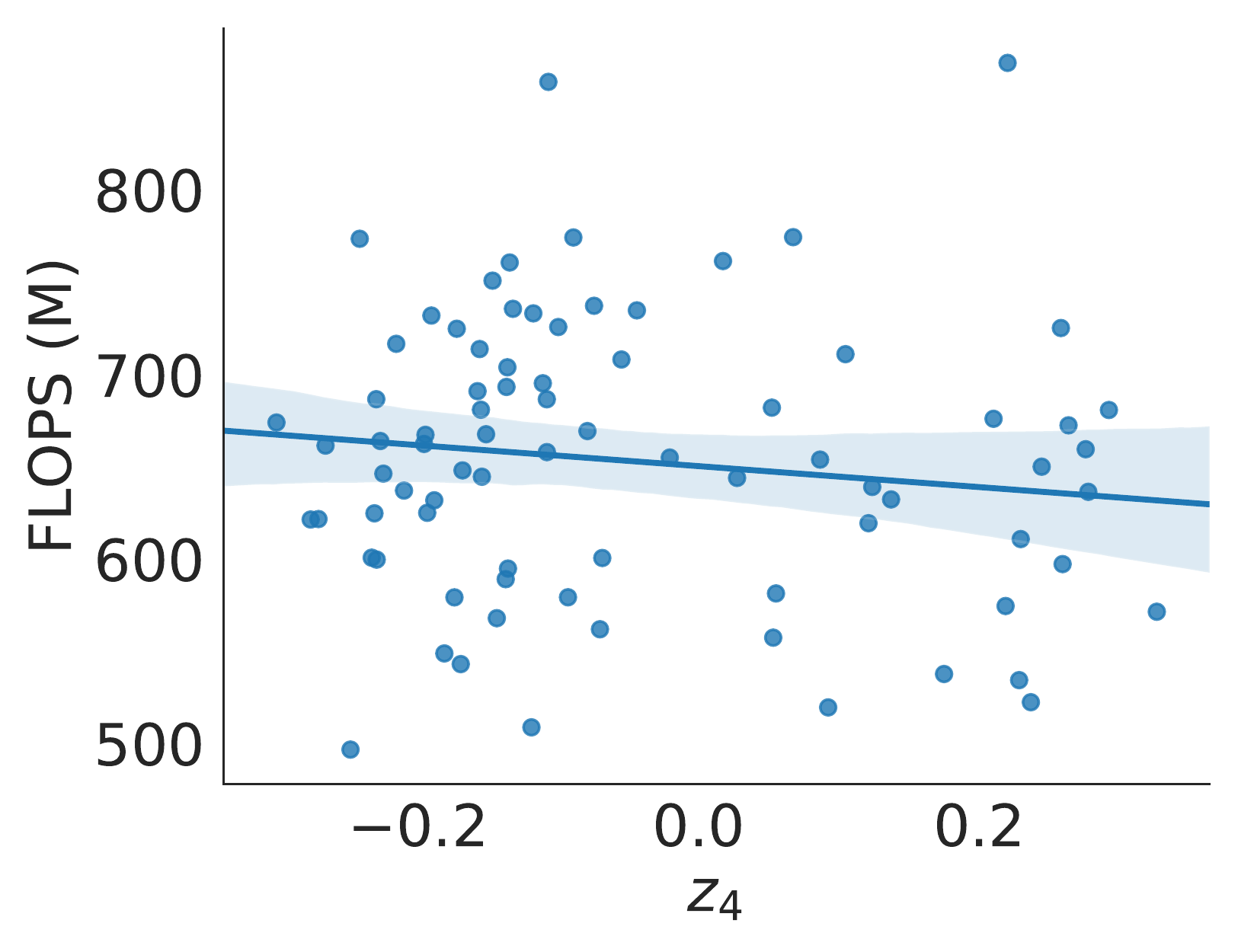}
\end{minipage}
}%
\hspace{.02in}
\subfigure[]{
\begin{minipage}[t]{0.23\linewidth}
\centering
\includegraphics[width =  1\linewidth]{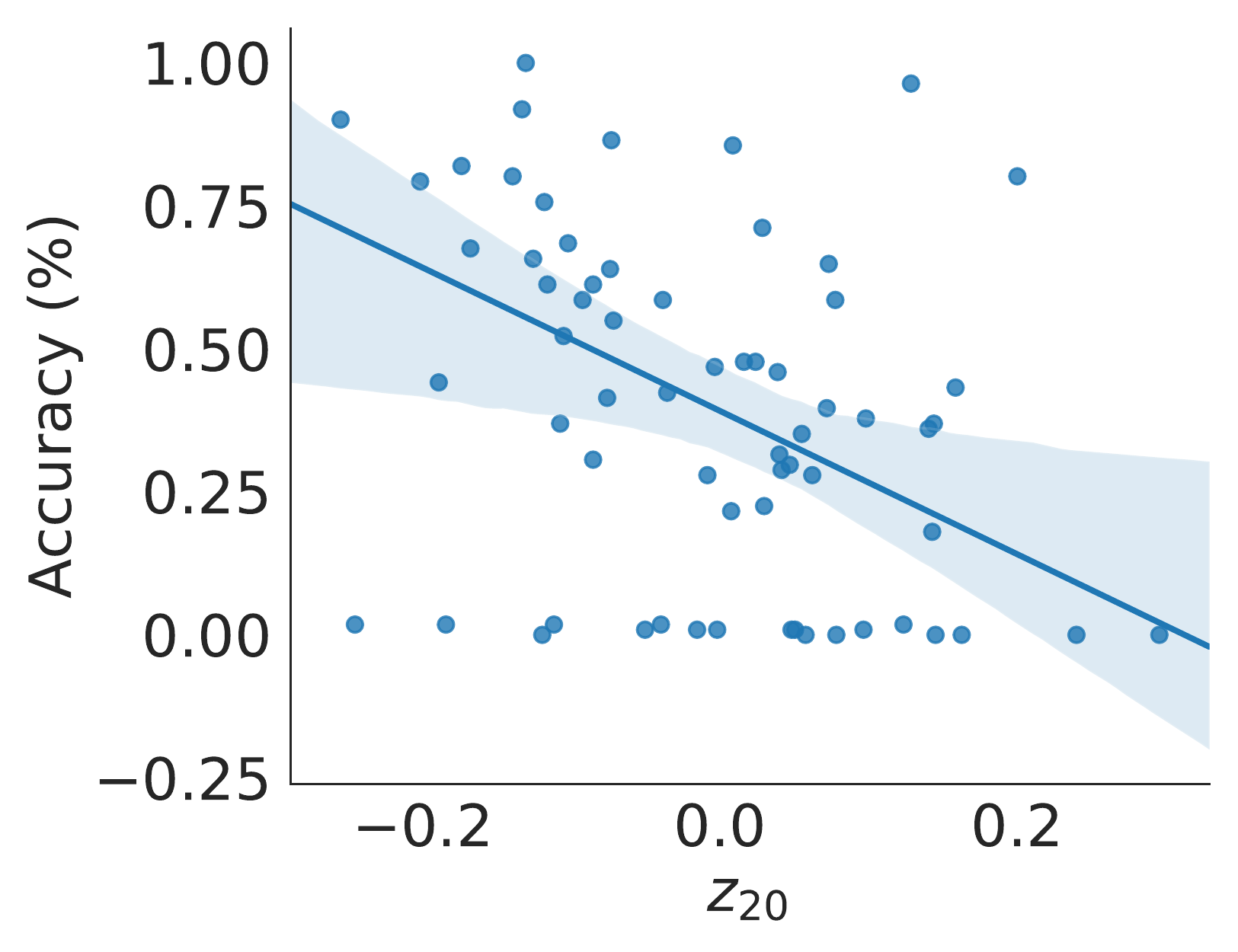}
\end{minipage}
}%
\hspace{.02in}
\subfigure[]{
\begin{minipage}[t]{0.23\linewidth}
\centering
\includegraphics[width =  1\linewidth]{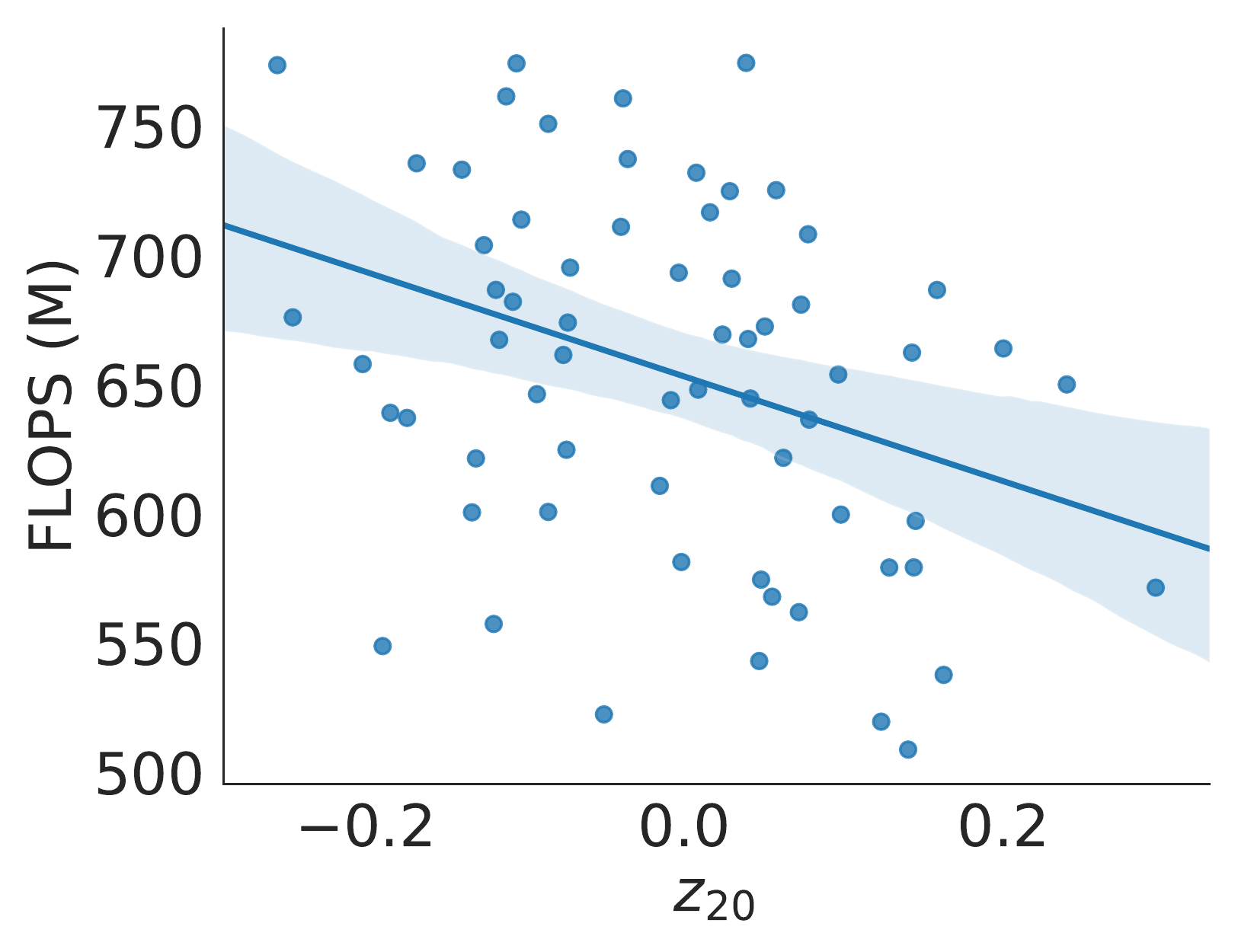}
\end{minipage}
}%

\hspace{.02in}
\subfigure[]{
\begin{minipage}[t]{0.35\linewidth}
\centering
\includegraphics[width =  1\linewidth]{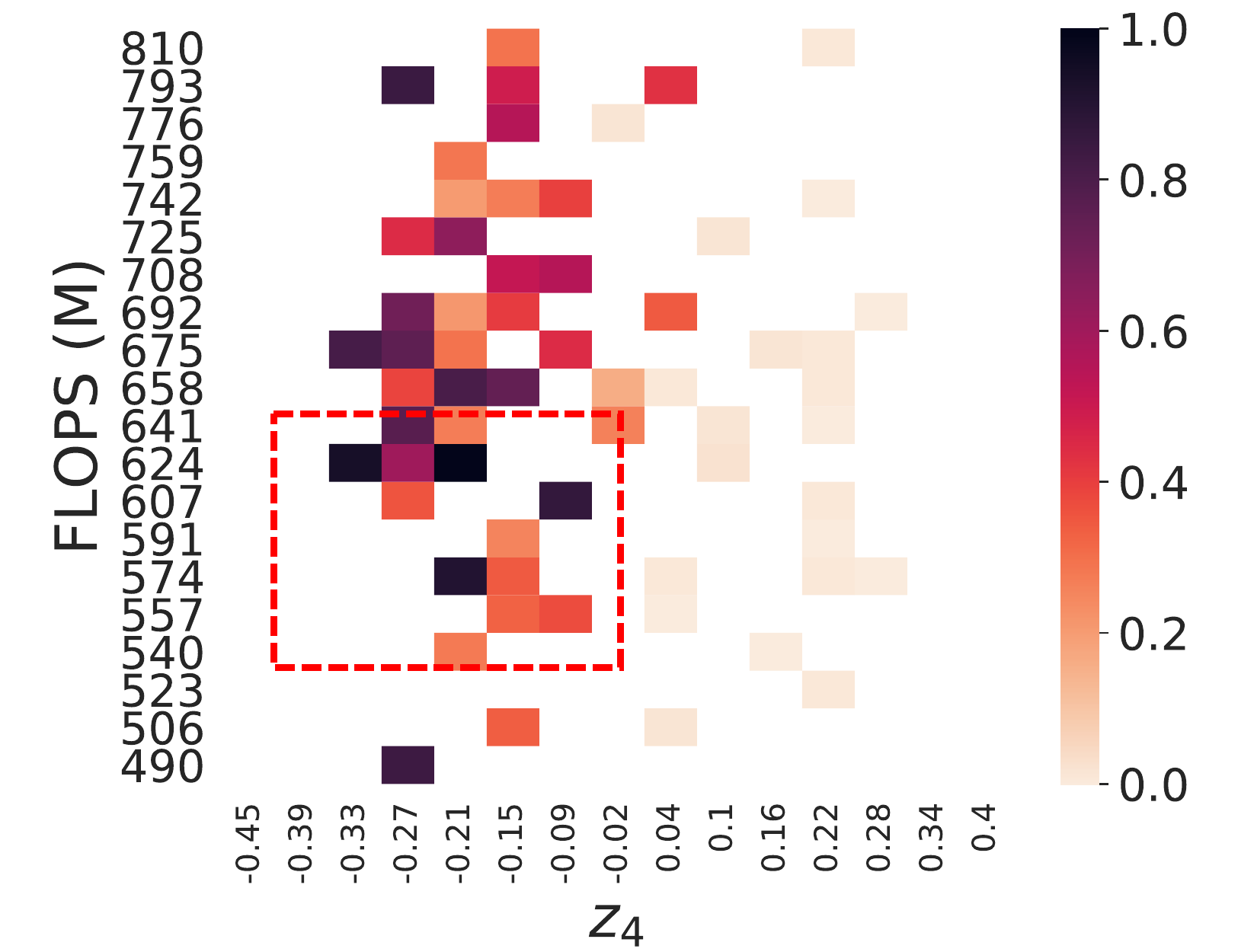}
\end{minipage}
}%
\hspace{.3in}
\subfigure[]{
\begin{minipage}[t]{0.35\linewidth}
\centering
\includegraphics[width =  1\linewidth]{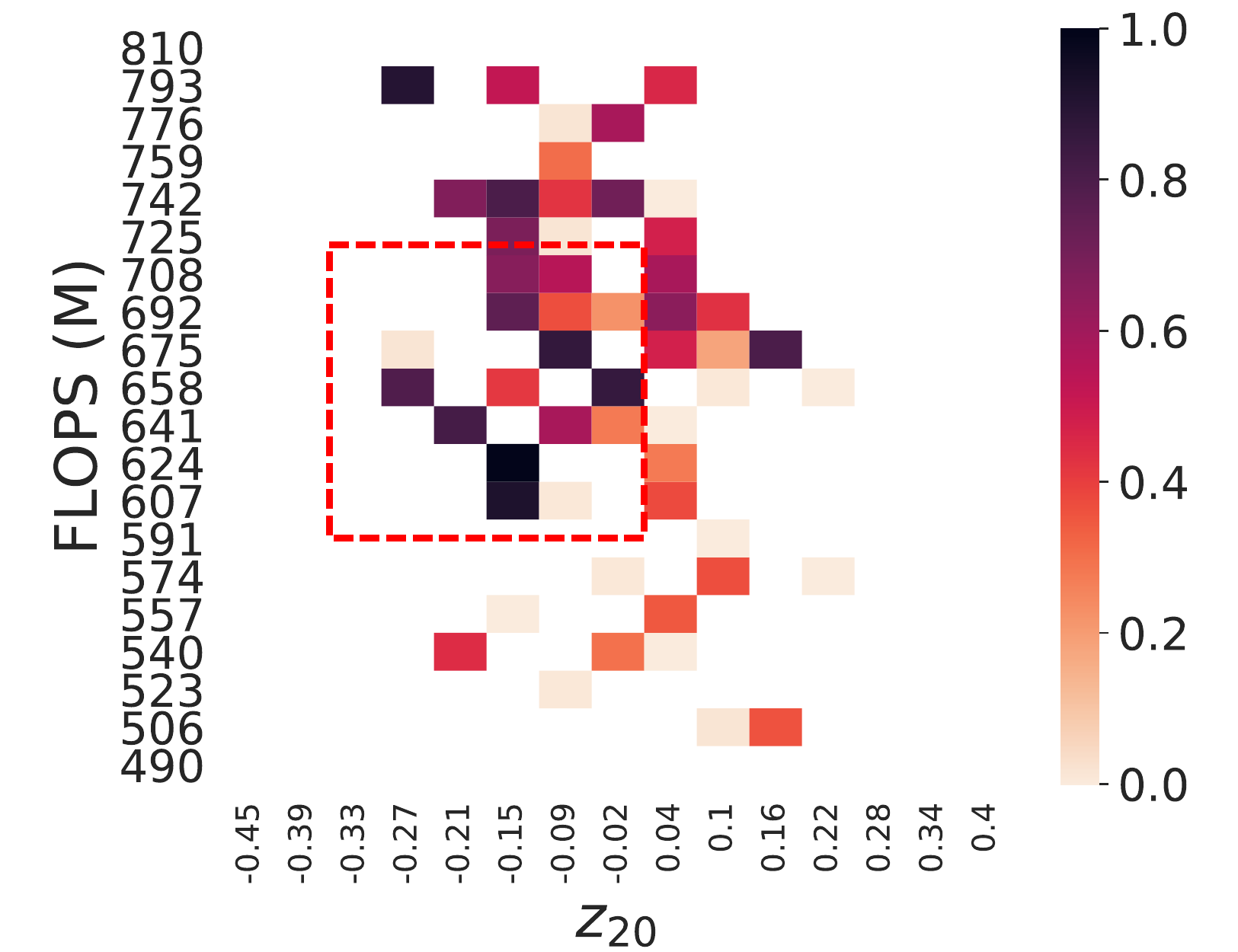}
\end{minipage}
}%

\hspace{.1in}
\centering
\caption{Correlation between disentangled factors, architecture performance and FLOPS on ImageNet. For $z_4$, (a) and (b) show there is no obvious relation between FLOPS and accuracy, so the idea sample region is within red frame in (e) where can reach highest performance with lower FLOPS (darker color presents higher accuracy). Similarly, in (c), (d) and (f), accuracy and FLOPS have a positive relation, so the sample area may locate slightly to upper left to compromise accuracy and FLOPS.}
\end{figure*}


\subsection{Dense-Sampling for Fast NAS}

\subsubsection{NASBench-101}

As shown in the previous section, the well-disentangled factors help us decompose the semantic concepts that independently control the representation of network architecture, which enables us to avoid the interference with other factors and intuitively determine the relation between the specific structural representation and performance.

We follow the Eqn 4. to determine the promising regions of dense-sampling. We set $\sigma$ = $0.05$ and take the $z$-value regions of the top-3 architectures for dense-sampling. For example, in Figure 4 (a), the $z_7$ values of the top-$3$ accuracy are $-0.28$, $-0.16$, and $-0.19$, so the promising regions are $[-0.33, -0.23]$ $\cup$ $[-0.21, -0.11]$ $\cup$ $[-0.24, -0.14]$, which result in $[-0.33, -0.11]$. Finally, we use the probability of $5\%$ to perform dense-sampling in the promising region and $95\%$ to sample in the entire search space. Since the search space of NASBench-101 is relatively simple, the composition of different architectures does not differ greatly in FLOPS. The resource restrictions are not considered here.

In Table 1, "Queries" indicates how many accuracies of architectures have been queried from NASBench-101, which is equivalent to evaluating the architectures. Therefore, reducing the number of queries can be considered as reducing computational cost. DNAS achieves $94.21\%$ best performance with the same number of queries as SemiNAS (accuracy $93.98\%$ is the top $0.01\%$ architecture in NASBench-101), and outperforms NAO and regularized evolution using less than $1/13$ computational cost (150 queries vs. 2000 queries).

\begin{table}[!h]\small
\renewcommand\arraystretch{1.2} 
\renewcommand\tabcolsep{10.0pt}
\centering  
\begin{tabular}{lccccc}  
\hline
\hline
Method & Queries & Accuracy(\%)  \\ 
\hline
Random Search   & $2000$ &$93.66$  \\
RE \cite{2018Regularized}   & $2000$ &$93.97$  \\
NAO \cite{NAO} & $2000$ & $93.87$ \\
\hline
SemiNAS \cite{semiNAS}   & $2100$ &$94.09$ \\
SemiNAS    & $300$ &$93.98$ \\

\hline

DNAS  &$300$ & $\textbf{94.21}$  \\
DNAS   & $150$ &$ 94.02 $ \\

\hline
\hline
\end{tabular}
\caption{Performance of different NAS algorithms on NASBench-101 dataset.}
\end{table}

\begin{table*}[!tp]\small
\renewcommand\arraystretch{1.2} 
\renewcommand\tabcolsep{5.0pt}
\centering  
\begin{tabular}{lccccc}  
\hline
\hline

Method & Top-1 Acc.(\%) & Top-5 Acc.(\%)  & Params(Million) & Flop(Million) \\
\hline

\textbf{DNAS} & $\textbf{23.4}$ &$\textbf{6.7}$  & $\textbf{6.20}$ & $\textbf{628}$  \\

SemiNAS(our impl.)  & $24.0$ &$7.1$  & $6.11$ & $611$  \\

\hline

\textbf{DNAS} & $\textbf{24.0}$ &$\textbf{6.9}$  & $\textbf{6.08}$ & $\textbf{569}$  \\

SemiNAS(our impl.) & $24.6$ &$7.4$  & $4.80$ & $580$  \\

Single-path NAS \cite{2020Single} & $25.0$ &$7.8$  & $-$ & $-$  \\

PC-DARTS \cite{2019PC} & $24.2$ &$7.3$  & $5.30$ & $597$  \\
DARTS \cite{DARTS} & $26.9$ &$9.0$  & $4.90$ & $595$  \\

PNAS \cite{progDNAS} &$25.8$ &$8.1$  & $5.10$ & $588$  \\

\hline
\textbf{DNAS} & $\textbf{24.5}$ &$\textbf{7.2}$  & $\textbf{5.05}$ & $\textbf{518}$  \\

SemiNAS(our impl.) & $24.9$ &$7.4$  & $4.55$ & $504$  \\
ProxylessNAS \cite{2017SMASH}   & $24.9$ &$7.5$  & $7.12$ & $465$  \\
Single Path One-shot\cite{SinglePathOne-shot} & $25.3$ &$\text{-}$  & $\text{-}$ & $328$ \\
SNAS\cite{2018SNAS} & $27.3$ &$9.2$  & $4.30$ & $522$ \\

\hline
\hline
\end{tabular} 
\caption{Performance of different NAS algorithms on ImageNet dataset.}
\end{table*}

\subsubsection{ImageNet}

ImageNet dataset consists of 1.28M training images and 5K test images, which are categorized into 1,000 image classes. For ImageNet experiments, we use slightly larger model than NASBench-101, which controller is a single layer LSTM with $46$ hidden size, the accuracy predictor is a $3$-layer MLP with hidden sizes of $46, 64, 1$ respectively, and the FLOPS predictor is a $2$-layer MLP with hidden sizes of $46, 1$ respectively. The top-100 architectures need to meet the restriction of $FLOPS(x) \leq F + \tau, (x=1,2, ...,100) $, where $\tau$ is 20M, otherwise it will be postponed to the sub-optimal architecture until 100 satisfactory architectures are found. The trade off in Eqn 5 is $\alpha = 0.8$, $\lambda = 0.3$, $\mu = 0.2$, and $\beta = 1$. For final evaluation, we train the searched network architecture for 300 epochs on 4 GPUs.

Specifically, we train the weight-sharing supernet for 60000 steps on 4 GPUs, then fix its well-trained parameters and sample the network structure from supernet for evaluation to train the controller. Since a well-trained supernet can guarantee a more accurate evaluation, the promising regions for dense-sampling will be more credible. Furthermore, the well-trained supernet can be saved for quick implementation of DNAS method under different FLOPS constraints.

Table 2 shows the performance of our DNAS method across 3 levels of resource limits (500M $\leq$ FLOPS $<$ 550M, 550M $\leq$ FLOPS $<$ 600M, 600M $\leq$ FLOPS). Compared to other baseline methods, DNAS achieves higher accuracy under the same computational constraints. Specifically, it outperforms the previous state-of-the-art method SemiNAS algorithm by 0.6\% and 0.5\% under mobile setting, indicating that the disentangling and dense-sampling methods work effectively in neural architecture search.

\begin{figure}[!h]
\centering
\includegraphics[scale=0.40]{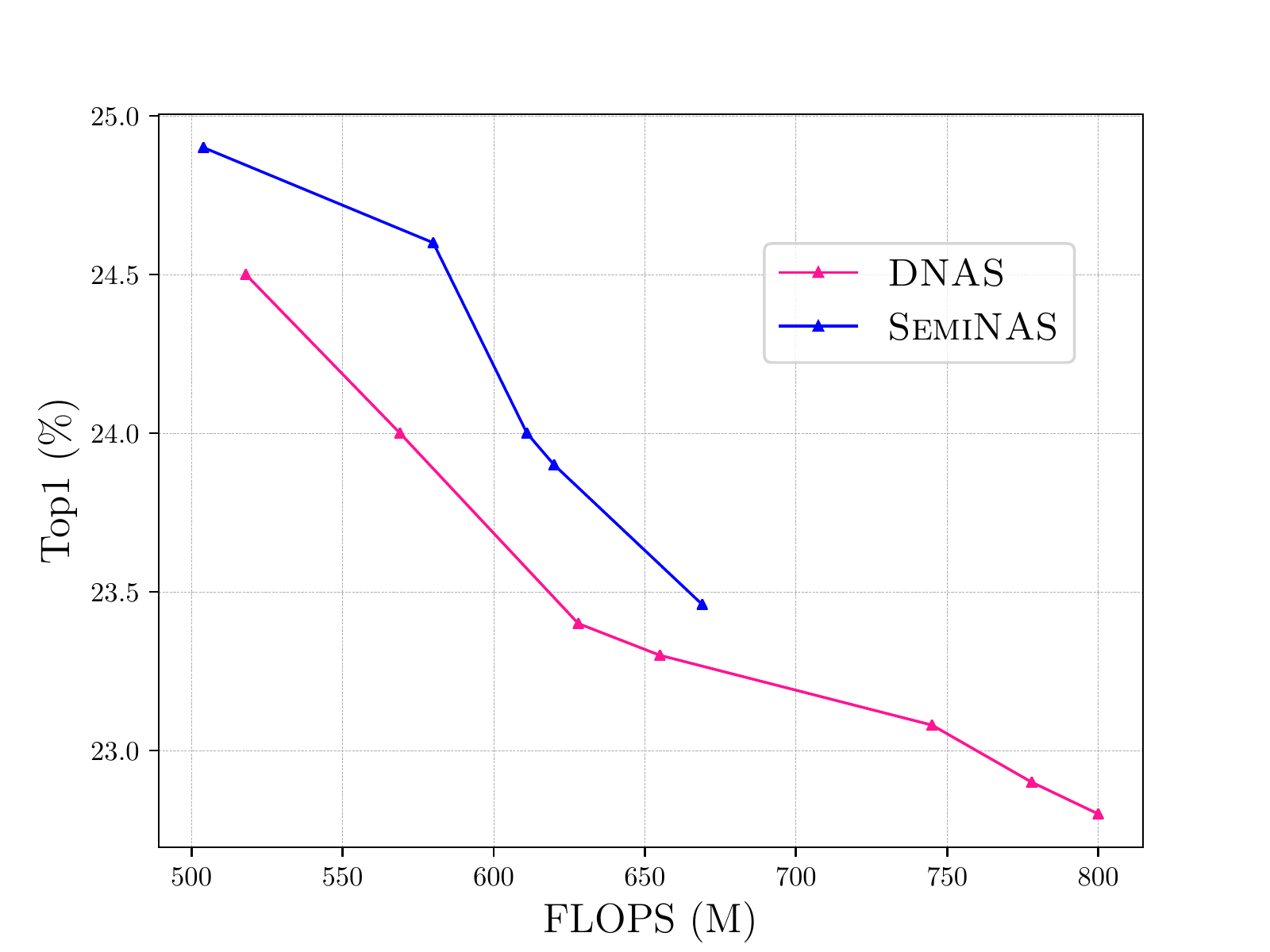}
\caption{Performance of SemiNAS VS. DNAS under different resource constraints.}
\end{figure}

When the FLOPS limit was increased to 700M or 800M, we run SemiNAS for 3 times, the discovered models are still distributed around 600M+ with no significant improvement in accuracy. In contrast, our DNAS method can find qualified and high-accuracy models (Figure 6). In practical engineering, DNAS will be more competent for given tasks.

\subsection{Ablation Study}

We conduct an ablation study to examine the effects of disentanglement and dense-sampling. DNAS queries $200$ architectures from NASBench-101. 

Two conclusions can be drawn from Table 3. First, compared to full model, all DNAS variants with some steps removed result in performance degradation, suggesting that both disentanglement and dense-sampling contributes to the success of DNAS. Second, the combination of disentangling and dense-sampling significantly improves performance, indicating that the disentangling procedure encourages the independence among latent factors. With help of that, we could easily identify which factors are more relevant to the accuracy and further perform dense-sampling effectively, which is usually extremely difficult for entangled representations.

\begin{table}[!h]\small
\renewcommand\arraystretch{1.2} 
\renewcommand\tabcolsep{5.0pt}
\centering  

\begin{tabular}{cccccc}  
\hline
\hline
Method & Dense & Disen. & Top-1 Acc.(\%) & Top-10 Avg Acc.(\%) \\ 
\hline
 \multirow{4}*{DNAS}  & $N$ &$N$  & $93.94 \pm 0.13$ & $93.59 \pm 0.11 $  \\
                        & $N$ &$Y$  & $93.94 \pm 0.09$ & $93.62 \pm 0.12 $  \\
                        & $Y$ &$N$  & $93.97 \pm 0.09$ & $93.65 \pm 0.08$  \\
                        & $Y$ &$Y$  & $94.05 \pm 0.12$ & $93.69 \pm 0.06$  \\

\hline
\hline

\end{tabular} 
\caption{The effect of disentangling and dense-sampling. We report the mean and std of accuracy over 20 runs. Disen. presents disentangling. Top-1 Acc. is the performance of the optimal architecture. Top-10 Avg Acc. is the average performance of discovered Top-10 architectures.}
\end{table}

\section{Conclusion}

In this work, we present our efforts towards interpretable and efficient neural architecture search. DNAS suggests to learn the disentangled representations of network architecture. Through analyzing the relation between the disentangled semantic representations and their performance, we perform dense-sampling in the semantic areas that have high possibility of generating good architectures. We demonstrate its consistently impressive performance and efficiency on NASBench-101 and ImageNet. Benefiting from disentanglement, we show that 1) the observed interpretable factors, including operation conversion and skip connection selection, provide a better understanding of neural networks design. 2) DNAS can perform dense-sampling at the edge of FLOPS constraints which makes full use of user resources to improve architecture performance. 3) Dense-sampling enables the controller to search well-performing architectures efficiently. We believe this work will lay the foundation for the interpretable neural architecture search, and contribute to further meaningful research.

\bibliography{reference}

\end{document}